\documentclass[sigconf]{acmart}

\usepackage{tabularx}
\usepackage{array}
\usepackage{makecell}

\AtBeginDocument{%
  }

\setcopyright{none}
\renewcommand\footnotetextcopyrightpermission[1]{}

\begin{document}

\title{TraceSQL: Traceable Answerability Estimation for Reference-Free Text-to-SQL Verification}



\author{Neelesh Kumar Shukla}
\authornote{Equal contribution.}
\affiliation{%
  \institution{Oracle Corporation}
  \country{India}}

\author{Debasmita Panda}
\authornotemark[1]
\affiliation{%
  \institution{Oracle Corporation}
  \country{India}}

\author{Srutanik Bhaduri}
\affiliation{%
  \institution{Oracle Corporation}
  \country{India}}

\author{Aditya Banerjee}
\affiliation{%
  \institution{Oracle Corporation}
  \country{United States}}

\author{Vasu Rangarajan}
\affiliation{%
  \institution{Oracle Corporation}
  \country{United States}}

\author{Viji Krishnamurthy}
\affiliation{%
  \institution{Oracle Corporation}
  \country{United States}}

\begin{abstract}

Text-to-SQL systems are commonly evaluated using ground-truth SQL queries or reference execution results, but such supervision is unavailable at inference time in real-world deployments. This creates a critical verification problem: given only a user question, database context, and generated SQL, can a system estimate whether the generated query is likely to correctly answer the question? Recent approaches use LLMs as judge or specialized agents to inspect generated SQL, but their decisions can be difficult to trace. Outcome Reward Models (ORMs) address this by learning from execution-labeled candidate SQLs and assigning correctness scores to unseen queries, yet they still provide limited visibility into the signals behind each verification. To address this limitation, we propose \textbf{TraceSQL}, a lightweight and traceable verification model built on explicit diagnostic features. TraceSQL combines 67 features capturing question ambiguity, question requirements, question--schema--SQL consistency, SQL structure, and intent alignment. These signals remain available for examining which factors influence each prediction and for tracing decisions back to diagnostic evidence. On BIRD development databases, TraceSQL achieves 66.47\% F1 and 64.48\% ROC-AUC, compared with 61.87\% F1 and 58.26\% ROC-AUC for the GradeSQL-7B ORM baseline on the same generated-SQL evaluation. Feature attribution further shows that the model relies on both semantic grounding and deterministic SQL-structure signals. These results show that SQL verification can be performed with a lightweight learned model while retaining feature-level evidence for inspecting and diagnosing its predictions.

\end{abstract}




\maketitle
\pagestyle{plain}
\raggedbottom

\section{Introduction}

Text-to-SQL systems have advanced rapidly with large language models (LLMs).
Benchmarks such as Spider~\cite{yu2018spider} and BIRD~\cite{li2023bird} evaluate systems on
complex questions and unseen database schemas. Recent methods further improve
generation through decomposition, example selection, self-correction,
multi-agent reasoning, and large-scale synthetic supervision~\cite{pourreza2023dinsql,gao2023dailsql,wang2025macsql,li2025omnisql}.
Despite these advances, generated SQL can contain subtle semantic,
schema-grounding, or structural errors. Reliable Text-to-SQL therefore requires not only generating a query, but also determining whether the generated query can be trusted. During benchmark evaluation, this decision can be supported by ground-truth SQL
or reference execution results. In deployment, however, such references are
typically unavailable, so post-generation verification must rely on the user
question, database context, generated SQL, and signals derived from their
relationships.
Existing approaches provide post-generation reliability in different ways. LLM-based methods use self-correction or specialized agents to inspect and refine generated SQL ~\cite{pourreza2023dinsql,wang2025macsql,askari2025magic}. More recent work has also explored reasoning-based
SQL judges and explicit rule-based verification
~\cite{bai2025judgesql,tian2026pvsql}. These approaches showcase different forms of verification evidence, such as correction reasoning, agent feedback, execution signals, and rule-based checks. However, this evidence is usually tied to the design of the individual method and is not represented in a unified, measurable form that can be systematically analyzed or linked to a learned verification decision. 

Outcome Reward Models (ORMs) provide a learned alternative to these approaches. GradeSQL~\cite{tritto2026gradesql} generates multiple candidate SQLs during
training, derives correctness labels through execution match with the reference
SQL, and fine-tunes an LLM to estimate candidate correctness. At inference time, the trained ORM assigns a probability-based correctness score without requiring the reference SQL. This demonstrates that
execution-derived supervision can be used to train a dedicated Text-to-SQL verifier. However, the resulting prediction is primarily exposed as a scalar score. While useful for ranking or selecting candidates, this score provides limited visibility into the specific semantic or structural signals that contributed to the decision.

This leaves an important gap when verification is expected to support not only prediction, but also understanding and diagnosis. A candidate may fail because of incorrect schema grounding, unsupported analytical requirements, missing or inappropriate SQL operations, incorrect grouping, or a mismatch with the intended result. Knowing that a candidate received a low verification score
does not by itself distinguish among these cases. For a verifier to support system analysis, its decision should be \emph{interpretable} in terms of meaningful signals, \emph{explainable} in terms of how those signals influence
the prediction, and \emph{traceable} back to the evidence from which those signals were derived.

We introduce \textbf{TraceSQL}, a lightweight learned verifier designed around these three properties. TraceSQL represents each question--candidate pair through explicit diagnostic signals rather than relying only on a final
verifier score. The representation combines five complementary diagnostic families: question ambiguity, question planning, pair analysis between the question, schema, and generated SQL, deterministic SQL structure, and intent alignment. Together, they produce 67 compact features describing the requirements expressed by the question, the grounding and semantic consistency
of the candidate, the operations present in the SQL, and its alignment with the intended request.

The representation provides three levels of visibility into the verification process. First, the model inputs are named diagnostic features with explicit
semantic meaning, making the representation itself interpretable. Second, feature-importance methods can be applied to the learned model to determine which signals influence its predictions, providing model-level explainability. Third, the richer diagnostic outputs from which the features are derived are retained separately, allowing influential features to be traced back to their
supporting evidence. TraceSQL therefore connects the final verification decision to both measurable feature-level signals and their diagnostic provenance.

We train TraceSQL using generated candidate SQLs from the balanced BIRD ORM training data released with GradeSQL. We then evaluate TraceSQL and GradeSQL-7B on the same generated-candidate SQLs from held-out BIRD databases.
TraceSQL achieves 66.47\% F1 and 64.48\% ROC-AUC, compared with 61.87\% F1 and 58.26\% ROC-AUC for GradeSQL-7B. Beyond predictive performance, our feature analysis shows that the learned verifier draws on both semantic
diagnostic signals and deterministic SQL-structure properties, allowing its behavior to be examined in terms of concrete properties of the generated query.

\paragraph{Contributions.}
We make three main contributions. First, we introduce \textbf{TraceSQL}, a lightweight learned verifier that combines SQL verification with interpretability, explainability, and traceability. Second, we develop a
67-feature representation from five diagnostic families in which model inputs remain semantically meaningful and retain provenance to their underlying diagnostic evidence. Third, we evaluate TraceSQL against GradeSQL-7B on
identical held-out generated-candidate SQLs and analyze the semantic and structural signals that drive its predictions, showing how verification decisions can be connected to explicit feature-level evidence.

\section{Related Work}

\subsection{Text-to-SQL Generation}

Recent LLM-based Text-to-SQL methods improve different parts of the generation process. DIN-SQL decomposes generation into schema linking, query decomposition, SQL generation, and self-correction ~\cite{pourreza2023dinsql}. DAIL-SQL focuses on prompt construction and example selection for in-context learning~\cite{gao2023dailsql}. MAC-SQL uses a multi-agent framework with dedicated components for decomposition, context reduction, and SQL refinement~\cite{wang2025macsql}. OmniSQL addresses the
training-data bottleneck through large-scale  synthetic supervision for specialized Text-to-SQL models~\cite{li2025omnisql}. These methods mainly
improve SQL generation or refinement, whereas TraceSQL operates after generation and focuses on verifying a produced candidate while preserving the evidence behind the verification decision.

\subsection{Post-Generation Verification and Learned Verifiers}

Recent work has increasingly focused on detecting and correcting errors after
SQL generation. MAGIC uses specialized agents to derive self-correction guidelines from generation failures~\cite{askari2025magic}, while DPC performs training-free candidate verification by comparing SQL behavior with an independently constructed execution path~\cite{li2026dpc}. Learned verification has also
emerged as an alternative: STaR-SQL incorporates an outcome-supervised reward model to rank generated candidates during test-time reasoning ~\cite{he2025starsql}.

GradeSQL~\cite{tritto2026gradesql} develops this direction further through
task-specific Outcome Reward Models (ORMs), trained on execution-derived
candidate labels and used to assign continuous correctness scores at inference
time. TraceSQL follows the idea of learning from execution-derived candidate labels, but differs in how verification is represented. Rather than
exposing only a scalar correctness score, TraceSQL retains a fixed set of diagnostic signals that can be analyzed individually and traced back to their
underlying evidence.

\subsection{Interpretable, Explainable, and Traceable Verification}

Recent verification methods provide different forms of transparency, including correction guidelines, structured reasoning, and explicit verification constraints~\cite{askari2025magic,bai2025judgesql,tian2026pvsql}. TraceSQL takes a complementary approach by representing verification evidence as explicit, measurable features that can be directly related to the behavior of
the learned verifier.

TraceSQL provides transparency at three levels. \emph{Interpretability} comes from semantically defined diagnostic features used as model inputs. \emph{Explainability} is provided through feature-importance and attribution methods, including permutation importance and SHAP\\
~\cite{lundberg2017shap}, which reveal which signals influence the model's predictions.
\emph{Traceability} is preserved by linking these features back to the diagnostic evidence from which they were derived. Together, these properties allow the verifier to expose not only its prediction, but also the semantic and
structural signals associated with that decision.

\section{Task Formulation}

Let $q$ denote a natural-language question, $s$ the corresponding database
context, and $x$ a candidate SQL to be verified. The verification task is to
estimate whether $x$ correctly answers $q$ using only information available to
the verifier. In the primary setting, $x$ is a generated candidate SQL; the
same verification pipeline is also applied to the ground-truth SQL provided
with the BIRD development databases.

During training and evaluation, candidate correctness is determined using
execution match:
\begin{equation}
y = \mathbb{1}\!\left[\mathrm{Exec}(x)=\mathrm{Exec}(x^\star)\right],
\end{equation}
where $x^\star$ denotes the reference SQL and $y \in \{0,1\}$ is the
candidate-level correctness label. The reference SQL and the resulting
execution-match label are used only to construct offline supervision and are
not provided as inputs to TraceSQL.

Given $(q,s,x)$, the diagnostic pipeline constructs a 67-dimensional feature
representation
\begin{equation}
\mathbf{z}=f(q,s,x),
\end{equation}
where $\mathbf{z}$ captures explicit signals related to question ambiguity,
question requirements, question--schema--SQL consistency, SQL
structure, and intent alignment. TraceSQL then estimates
\begin{equation}
\hat{p}=P(y=1\mid\mathbf{z}),
\end{equation}
where $\hat{p}$ is the estimated probability that the candidate SQL is
correct.

Using a fixed decision threshold $\tau=0.50$, the final verification decision is
\begin{equation}
\hat{y}=\mathbb{1}[\hat{p}\geq\tau].
\end{equation}

The features in $\mathbf{z}$ remain individually inspectable and retain their
diagnostic provenance. This allows the verification decision to be examined in
terms of the semantic, grounding, and structural signals that influence the
learned prediction.
\section{Traceable Answerability Signals}

\subsection{Diagnostic Evidence Generation}

TraceSQL derives its verification evidence from three upstream diagnostic
modules: the Ambiguity Detector, Pair Analyzer, and SQL Repair Module. Together,
they capture question-level ambiguity, question--schema--SQL consistency, and
alignment between the user request and the candidate SQL. Their structured
outputs are retained as diagnostic evidence and subsequently transformed into
the model features described in Section~\ref{sec:feature_extraction}.

\subsubsection{Ambiguity Detector}

The Ambiguity Detector assesses whether the user question is sufficiently
specified with respect to the available database context. It produces:

\begin{itemize}
    \item \textbf{Ambiguity status}: indicates whether the question is
    considered ambiguous.

    \item \textbf{Ambiguity probability}: estimates the degree of ambiguity
    on a 0--100 scale.

    \item \textbf{Explanation}: provides the evidence or rationale supporting
    the ambiguity assessment.
\end{itemize}

These outputs provide the source evidence for constructing the downstream
ambiguity features.

\subsubsection{Pair Analyzer}

The Pair Analyzer evaluates whether the candidate SQL is supported by the user
question and database context. It applies predefined verification rules covering
schema and column grounding, joins, identifiers, data types, temporal semantics,
metrics, null and validity logic, and business-term mappings. Its diagnostic
outputs include:

\begin{itemize}
    \item \textbf{Support status}: classifies the candidate as
    \textit{supported}, \textit{partially supported}, \textit{unsupported},
    or \textit{unclear}.

    \item \textbf{Pair score and summary}: provide an overall assessment of
    candidate support.

    \item \textbf{Rule-level results}: record the status, score, explanation,
    and supporting evidence for each verification rule.

    \item \textbf{Findings}: describe detected grounding or consistency gaps
    and the affected database or SQL elements.

\end{itemize}

The rule-level statuses are used to construct the Pair Analysis feature family,
while the richer explanations and findings are retained as diagnostic evidence
for traceability.

\subsubsection{SQL Repair Module}

The SQL Repair Module is used in \\
diagnosis-only 
mode to assess how well the
candidate SQL reflects the requirements expressed by the user question. It
decomposes the question into structured intents, explains the behavior of the
candidate SQL, and evaluates their alignment. The outputs retained for this
work are:

\begin{itemize}
    \item \textbf{Question intent breakdown}: decomposes the question into
    ordered, schema-grounded requirements that the SQL is expected to satisfy.

    \item \textbf{SQL explanation}: describes the operations performed by
    the candidate SQL, including tables, columns, joins, filters,
    aggregations, and other query constructs.

    \item \textbf{Reference-free evaluation}: assesses the candidate against
    the planned question intents and produces an evaluation score, confidence,
    gate decision, intent-level alignment assessments, and an intervention
    reason when a mismatch is detected.
\end{itemize}

These outputs provide the source evidence for constructing the downstream
question-planning and intent-alignment features.

\subsection{Feature Extraction}
\label{sec:feature_extraction}

Feature extraction depends on the type of diagnostic evidence. For the
ambiguity, question-planning, and intent-alignment families, TraceSQL uses
predefined probe sets encoded in fixed evaluation prompts. Relevant diagnostic
units---including ambiguity claims, question-intent items, and intent-alignment
evidence---are supplied to an LLM together with the required question, schema,
and SQL context. The LLM evaluates each unit against the applicable probes and
returns a structured verdict (\textit{PASS}, \textit{FAIL}, \textit{N/A}, or
\textit{UNKNOWN}). The LLM configurations used across the diagnostic pipeline
are reported in Section~\ref{sec:llm_config}.

\begin{figure*}[t]
    \centering
    \includegraphics[width=1.10\textwidth]
    {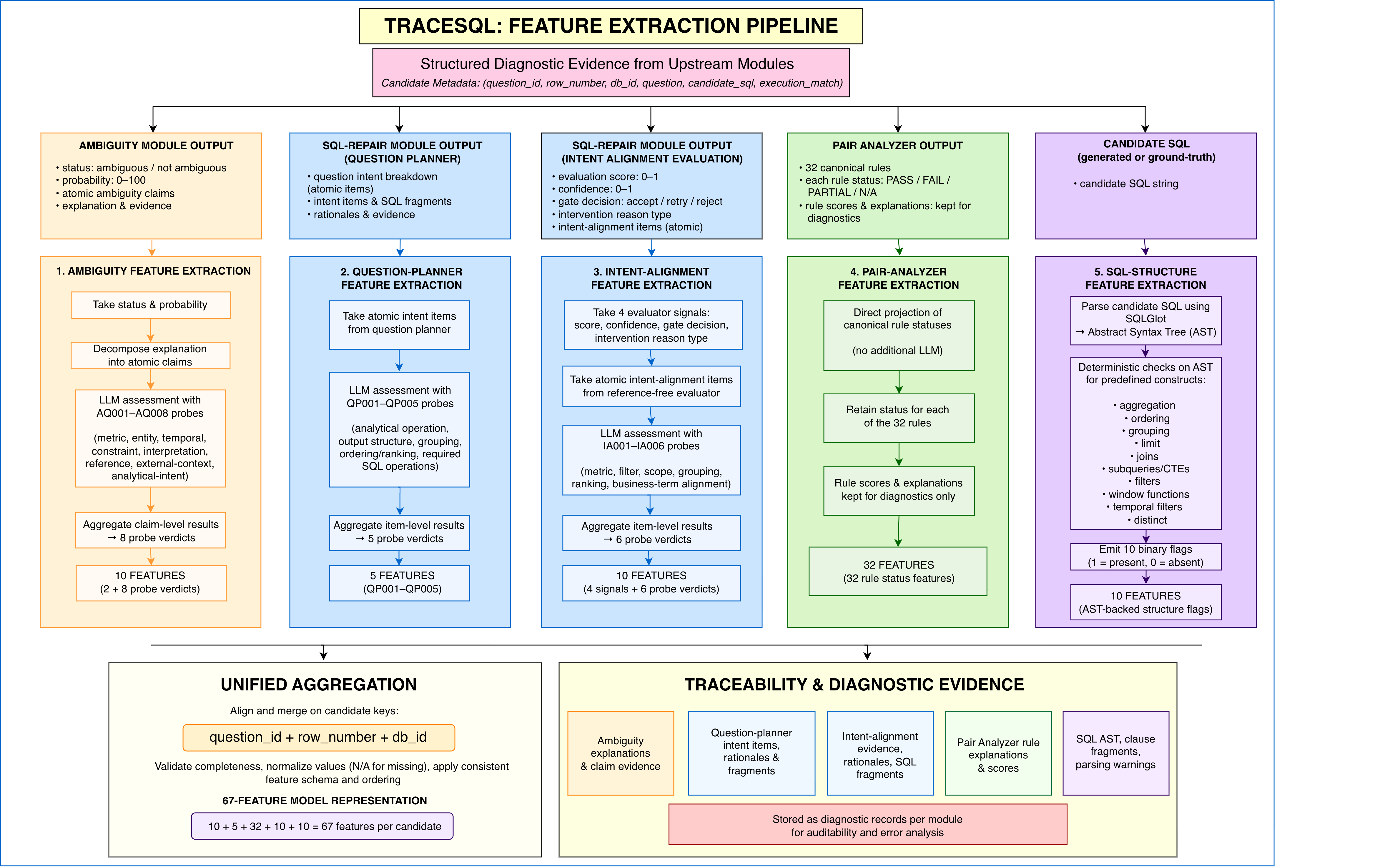}
    \caption{TraceSQL feature-construction pipeline. Diagnostic evidence is transformed into five feature families through LLM-based probe assessment, direct projection of Pair Analyzer rule statuses, and deterministic SQL-AST extraction using SQLGlot. The resulting $10+5+32+10+10=67$ features are aligned at the candidate level and merged into the unified model representation, while richer diagnostic evidence is retained separately for traceability.}
    \label{fig:feature_extraction_pipeline}
\end{figure*}

Pair Analysis requires no additional LLM assessment during feature
construction; the statuses of predefined canonical rules are projected directly
into the model representation. SQL-structure features are also deterministic:
the candidate SQL is parsed into an abstract syntax tree (AST) using
\texttt{SQLGlot}, from which predefined structural properties are extracted.
Figure~\ref{fig:feature_extraction_pipeline} summarizes the resulting
feature-construction pipeline. 

\subsubsection{Ambiguity Features}

The ambiguity feature pipeline transforms the Ambiguity Detector output into ten structured features. Ambiguity status and probability are retained directly, while the detector explanation is decomposed into claims and evaluated against eight predefined ambiguity probes covering metric, entity, temporal, constraint, interpretation, reference, external-context, and analytical-intent ambiguity. Claim-level assessments are aggregated into one verdict per probe,
yielding eight probe features in addition to the two detector-level
signals.

\subsubsection{Question-Planning Features}

The question-planning feature pipeline transforms the \textit{question intent breakdown} produced by the SQL Repair Module into five structured features.
Each atomic intent item is evaluated by an LLM against five predefined planning probes covering the analytical operation, expected output structure, grouping,
ordering or ranking, and required SQL operations. Item-level assessments are aggregated into one verdict per probe,
producing five candidate-level question-planning features. Supporting evidence, rationale, confidence, and planning fragments are
retained separately for traceability.

\subsubsection{Pair-Analysis Features}

The pair-analysis feature pipeline directly transforms the rule-level outputs of the Pair Analyzer into 32 structured features. Unlike the ambiguity and question-planning pipelines, no additional LLM assessment is performed during
this stage. Each predefined Pair Analyzer rule contributes one categorical feature corresponding to its rule status, while the associated rule score and explanation are retained separately as diagnostic evidence. The rules cover
schema and scope grounding, joins, identifiers, data types, temporal semantics, metric definitions, data validity, and business-term mappings. 

\subsubsection{SQL-Structure Features}

The SQL-structure pipeline extracts ten deterministic features directly from
the candidate SQL. Each query is parsed into an abstract syntax tree (AST)
using \texttt{SQLGlot}, which is inspected for aggregation, grouping, joins,
filtering, temporal filtering, ordering, limits, subqueries or CTEs, window
functions, and \texttt{DISTINCT}. Each property is represented as a binary
feature indicating its presence or absence. The parsed AST, clause-level SQL
evidence, and parsing warnings are retained separately for traceability.

\subsubsection{Intent-Alignment Features}

The intent-alignment feature \\
pipeline transforms the structured output of the reference-free evaluator into ten candidate-level features. Four evaluator signals---the overall evaluation score, confidence, gate decision, and
intervention-reason type---are retained directly. In addition, each intent-alignment item is evaluated by an LLM against six predefined probes covering metric, filter, scope, grouping, ranking, and business-term alignment.
Item-level assessments are aggregated into one verdict per probe, producing six additional features. Supporting evidence, rationales, SQL fragments, and the original intent-alignment records are retained separately for traceability.

Table~\ref{tab:feature_family_overview} summarizes the resulting
67-dimensional diagnostic representation and provides representative signals
from each feature family. Complete definitions of all 67 features, including
their probe and canonical-rule identifiers, are provided in the supplementary
material.

\begin{table*}[!t]
\centering
\caption{Overview of the five TraceSQL diagnostic feature families and
representative signals captured by each family.}
\label{tab:feature_family_overview}

\small
\setlength{\tabcolsep}{4.5pt}
\renewcommand{\arraystretch}{1.05}

\begin{tabularx}{\textwidth}{@{}l c l X@{}}
\toprule
\textbf{Feature Family} &
\textbf{\#} &
\textbf{Representative Features} &
\textbf{Captured Evidence} \\
\midrule

Ambiguity &
10 &
\makecell[l]{Ambiguity Overall Status;\\
Metric Clarity; Time-Scope Clarity} &
Whether the question is sufficiently specified with respect to the requested
metric, entities, temporal scope, constraints, references, and analytical
intent. \\

Question Planning &
5 &
\makecell[l]{Question Planner: Analytical Operation;\\
Question Planner: Output Structure;\\
Question Planner: Grouping} &
Whether the candidate SQL reflects the analytical operation, expected output
structure, grouping, ordering, and SQL operations required by the question. \\

Pair Analysis &
32 &
\makecell[l]{Join-Key Support;\\
Data-Validity Rule Support;\\
Additional Grounding Check} &
Whether tables, columns, joins, identifiers, data types, temporal semantics,
metrics, validity rules, and business concepts are grounded in the available
database context. \\

SQL Structure &
10 &
\makecell[l]{SQL Aggregation Usage;\\
SQL LIMIT Usage;\\
SQL DISTINCT Usage} &
Deterministic structural properties of the candidate SQL extracted from its
SQLGlot abstract syntax tree. \\

Intent Alignment &
10 &
\makecell[l]{Intent Alignment: Evaluation Score;\\
Intent Alignment: Grouping;\\
Intent Alignment: Business Term} &
Whether the candidate SQL aligns with the requested metric, filters, scope,
grouping, ranking, and business terminology. \\

\midrule
\textbf{Total} &
\textbf{67} &
&
\\
\bottomrule
\end{tabularx}
\end{table*}

\section{Research Questions}
\label{sec:research-questions}

We evaluate TraceSQL from four complementary perspectives: verification
performance, diagnostic-signal importance, cross-database generalization,
and traceability. We investigate the following research questions.

\begin{itemize}
    \item[\textit{RQ1:}] How effectively can structured diagnostic evidence
    support reference-free Text-to-SQL verification?

    \item[\textit{RQ2:}] Which diagnostic signals contribute most to the
    learned verification decision?

    \item[\textit{RQ3:}] How well does TraceSQL generalize across held-out
    databases?

    \item[\textit{RQ4:}] To what extent can TraceSQL predictions be traced
    back to explicit diagnostic evidence?
\end{itemize}

\section{Data Preparation}

We use the balanced BIRD ORM training corpus released with
GradeSQL~\cite{tritto2026gradesql}. After attaching database identifiers, the
corpus contains 30,686 candidate SQL queries from 69 databases and 2,642 unique
question--database groups. Each group contains between 2 and 32 candidates, so
sampling individual rows would disproportionately represent questions with
larger candidate sets.

Database identifiers are assigned by matching normalized \\
question-plus-evidence text to the corresponding BIRD training records. We then
remove only exact duplicates across question, schema, SQL, data, label, and
database ID. This reduces the corpus from 30,686 to 19,592 candidates while
preserving alternative candidate SQLs for the same question.

For each eligible question--database group, we retain one positive and one
negative candidate, yielding 2,642 balanced pairs, or 5,284 candidates. From
these, we select 1,000 complete pairs (2,000 candidates) using
database-stratified sampling. All 69 training databases remain represented,
with the remaining samples allocated proportionally across databases. The
resulting training set contains 1,000 positive and 1,000 negative candidates
and is reproducible using seed 42.

The final 2,000-candidate sample represents approximately 6.5\% of the original
ORM corpus and 37.9\% of the 5,284-candidate paired source. We use this sample for the current TraceSQL training experiments, while
scaling to the complete paired source is left for future work.

\section{Experimental Setup}

\subsection{LLM Configuration for Diagnostic Evidence and Feature Extraction}
\label{sec:llm_config}

The diagnostic pipeline uses fixed LLM backends for evidence generation and
LLM-based feature extraction. The Ambiguity Detector and SQL Repair Module use
\texttt{GPT-4o}, while the Pair Analyzer uses \texttt{GPT-5.2}. In the unified feature-extraction pipeline, the ambiguity,
question-planning, and intent-alignment probe assessments are performed using
\texttt{GPT-4o}.

\subsection{Training with FLAML}

We train TraceSQL using FLAML~\cite{wang2021flaml} with all 67 diagnostic
features as input. The training set contains 2,000 candidates, balanced between
1,000 positive and 1,000 negative \texttt{execution\_match} labels. Following
FLAML's preprocessing and feature transformation, 62 features are retained as
inputs to the selected estimator. Thus, TraceSQL is defined over the complete
67-feature diagnostic representation, while the fitted estimator operates on
the corresponding 62-feature transformed representation.

FLAML uses internal validation ROC-AUC for model selection and subsequently
refits the selected estimator on all 2,000 training candidates. Metrics computed
on these candidates therefore describe training-population performance rather
than held-out evaluation.

With a 1,800-second search budget, FLAML evaluates 4,228 trials and selects an
Extra Trees classifier~\cite{geurts2006extratrees} with seven trees,
entropy-based splitting, \texttt{max\_leaves}=6, and
\texttt{max\_features}=0.2241. Increasing the search budget to 3,600 seconds
returns the same model configuration and performance, indicating that the
search had stabilized within the shorter budget.

\subsection{Baseline and External Evaluation}

The main baseline is GradeSQL-7B~\cite{tritto2026gradesql}. For external
evaluation, the saved 1,800-second TraceSQL model is applied without refitting
to 11 BIRD development databases that are disjoint from the training databases.
We evaluate two complementary settings. The generated-SQL setting contains both
execution-matching and non-matching candidates and serves as the primary
two-class verification evaluation. The ground-truth SQL setting evaluates the ground-truth SQL provided with
the BIRD development databases and serves as a complementary assessment of
verifier behavior on these queries.

GradeSQL-7B and TraceSQL are evaluated on identical generated-candidate SQLs
with the same execution-match labels. This provides a matched evaluation of the
two verification approaches on the same test instances.

\section{Results}

\subsection{Training Results}

The selected Extra Trees model achieves an internal validation ROC-AUC of 0.6109.
After refitting on all 2,000 training candidates, it reaches an accuracy of
0.6085, precision of 0.5987, recall of 0.6580, F1 of 0.6270, and ROC-AUC of
0.6520. These values characterize model development and training-population
performance rather than held-out evaluation.

\subsection{Which Diagnostic Signals Does the Model Use?}

We analyze the fitted 1,800-second model using native Extra Trees feature importance,
permutation importance, and exact Tree SHAP\\
~\cite{lundberg2017shap}. We use
these methods to characterize which diagnostic signals the learned verifier
relies on. Permutation importance and SHAP provide complementary
perturbation-based and attribution-based views, while native Extra Trees
importance is used as a supporting analysis.

The leading signals span both deterministic SQL structure and semantic
diagnostic evidence. Permutation importance ranks
\emph{SQL DISTINCT Usage}, \emph{SQL LIMIT Usage},
\emph{SQL Aggregation Usage}, \emph{Additional Grounding Check},
\emph{Pattern-Based Business Meaning Support}, and
\emph{Data-Validity Rule Support} among the strongest features. SHAP
independently highlights the same broad set. Nine features appear within the top ten under both permutation importance and
SHAP: \emph{Additional Grounding Check}, \emph{SQL LIMIT Usage},
\emph{Pattern-Based Business Meaning Support},
\emph{SQL DISTINCT Usage}, \emph{SQL Aggregation Usage},
\emph{Data-Validity Rule Support}, \emph{SQL Grouping Usage},
\emph{Intent Alignment: Grouping}, and
\emph{Question Planner: Output Structure}. This agreement shows that the
model relies on a combination of deterministic SQL-structure signals and
semantic diagnostic evidence rather than on a single type of signal.

Table~\ref{tab:importance} reports the six highest-ranked features by
permutation importance together with their mean absolute SHAP values.

The strongest signals are distributed across multiple parts of the TraceSQL
representation. SQL-structure features such as \emph{SQL DISTINCT Usage},
\emph{SQL LIMIT Usage}, and \emph{SQL Aggregation Usage} appear alongside
grounding and semantic-validity signals from the Pair Analysis family. The fitted verifier
therefore combines evidence about the operations present in the candidate SQL
with evidence about whether those operations are supported by the question
and database context.

A more detailed feature-level analysis, including global SHAP analysis,
cross-method comparison of SHAP, permutation importance, and native Extra
Trees importance, feature correlation analysis, and a representative
prediction trace, is provided in the supplementary material.
\begin{table}[!tb]
\caption{Leading features by permutation importance in the selected TraceSQL
model. Permutation values are ROC-AUC decreases after shuffling; SHAP values
are mean absolute contributions.}
\label{tab:importance}
\centering
\small

\begin{tabular}{lrr}
\toprule
Feature & Perm. & Mean $|\mathrm{SHAP}|$ \\
\midrule
SQL DISTINCT Usage
    & 0.0344 & 0.0074 \\
SQL LIMIT Usage
    & 0.0335 & 0.0147 \\
SQL Aggregation Usage
    & 0.0327 & 0.0073 \\
Additional Grounding Check
    & 0.0317 & 0.0199 \\
Pattern-Based Business Meaning Support
    & 0.0219 & 0.0117 \\
Data-Validity Rule Support
    & 0.0165 & 0.0070 \\
\bottomrule
\end{tabular}
\end{table}

\subsection{Generated-SQL Verification and Cross-Database Generalization}
\label{subsec:generated_sql_eval}

We evaluate TraceSQL and GradeSQL-7B on 1,521 generated SQLs from
11 held-out BIRD development databases. As shown in
Table~\ref{tab:generated-overall}, TraceSQL outperforms GradeSQL-7B across all
aggregate metrics, with improvements of 4.60 percentage points in F1 and
6.22 points in ROC-AUC. These results indicate that the structured diagnostic
representation supports effective verification on databases not observed
during training.

\begin{table}[!tb]
\caption{Overall generated-SQL performance on 1,521 candidates from 11 held-out
BIRD development databases. Bold indicates the better result for each metric.}
\label{tab:generated-overall}
\centering
\small

\begin{tabular}{lccccc}
\toprule
Model & Acc. & Prec. & Rec. & F1 & AUC \\
\midrule
GradeSQL-7B
& 57.46
& 63.64
& 60.21
& 61.87
& 58.26 \\

TraceSQL
& \textbf{62.46}
& \textbf{68.11}
& \textbf{64.91}
& \textbf{66.47}
& \textbf{64.48} \\
\bottomrule
\end{tabular}
\end{table}

Table~\ref{tab:generated-db} shows that TraceSQL's aggregate gains are reflected
across a broad range of held-out databases. TraceSQL achieves stronger results
on most reported metrics for nine of the eleven databases, including
\emph{California Schools}, \emph{Codebase Community}, \emph{Financial},
\emph{Superhero}, and \emph{Toxicology}. Particularly large improvements are
observed for databases such as \emph{Superhero} and
\emph{Thrombosis Prediction}. Performance is more competitive on
\emph{Debit Card Specializing}, while \emph{Formula 1} is the main case where
GradeSQL-7B remains stronger on most metrics. Overall, the database-level
results suggest that the TraceSQL representation transfers effectively across
diverse schemas and query distributions, while leaving room for further
improvement on specific domains.

\begin{table*}[!t]
\centering
\caption{Database-level generated-SQL performance. Bold indicates the better
result for each metric within a database.}
\label{tab:generated-db}

\small
\setlength{\tabcolsep}{3.2pt}
\renewcommand{\arraystretch}{1.05}

\begin{tabular}{@{}lcccccccccc@{}}
\toprule
Database &
\multicolumn{5}{c}{GradeSQL-7B} &
\multicolumn{5}{c}{TraceSQL} \\
\cmidrule(lr){2-6}
\cmidrule(lr){7-11}
& Acc. & Prec. & Rec. & F1 & AUC
& Acc. & Prec. & Rec. & F1 & AUC \\
\midrule

California Schools
& 45.45 & 64.86 & 40.68 & 50.00 & \textbf{58.33}
& \textbf{51.14} & \textbf{69.05} & \textbf{49.15}
& \textbf{57.43} & 56.25 \\

Card Games
& 55.26 & 50.47 & \textbf{62.79} & 55.96 & 53.98
& \textbf{61.05} & \textbf{56.82} & 58.14
& \textbf{57.47} & \textbf{64.73} \\

Codebase Community
& 61.83 & 73.45 & 66.94 & 70.04 & 59.31
& \textbf{65.59} & \textbf{75.00} & \textbf{72.58}
& \textbf{73.77} & \textbf{66.03} \\

Debit Card Specializing
& 48.33 & 44.19 & \textbf{73.08} & \textbf{55.07} & 46.27
& \textbf{60.00} & \textbf{55.00} & 42.31
& 47.83 & \textbf{60.35} \\

European Football 2
& 57.14 & 71.43 & 55.56 & 62.50 & 59.48
& \textbf{61.90} & \textbf{75.38} & \textbf{60.49}
& \textbf{67.12} & \textbf{69.36} \\

Financial
& 57.55 & 56.92 & \textbf{68.52} & 62.18 & 56.91
& \textbf{64.15} & \textbf{63.79} & \textbf{68.52}
& \textbf{66.07} & \textbf{63.46} \\

Formula 1
& \textbf{61.27} & \textbf{65.98} & \textbf{65.31}
& \textbf{65.64} & 63.31
& 60.12 & 65.93 & 61.22
& 63.49 & \textbf{65.27} \\

Student Club
& 60.65 & 70.64 & 72.64 & 71.63 & 47.19
& \textbf{61.94} & \textbf{71.56} & \textbf{73.58}
& \textbf{72.56} & \textbf{57.20} \\

Superhero
& 54.26 & 73.49 & 62.24 & 67.40 & 36.87
& \textbf{68.99} & \textbf{81.52} & \textbf{76.53}
& \textbf{78.95} & \textbf{62.66} \\

Thrombosis Prediction
& 57.67 & 45.45 & 30.77 & 36.70 & 56.89
& \textbf{61.96} & \textbf{52.11} & \textbf{56.92}
& \textbf{54.41} & \textbf{60.76} \\

Toxicology
& 60.69 & 64.06 & 54.67 & 58.99 & 56.91
& \textbf{65.52} & \textbf{66.67} & \textbf{66.67}
& \textbf{66.67} & \textbf{67.61} \\

\bottomrule
\end{tabular}
\end{table*}

\subsection{Traceability of Verification Decisions}
TraceSQL preserves the connection between influential model signals and
explicit diagnostic evidence through the feature-extraction process: every
model feature is associated with a named probe, canonical verification rule,
or deterministic SQL-structure property, while the richer supporting evidence
is retained under the same candidate identity.

For example, SQL-structure features such as \emph{SQL LIMIT Usage}
can be traced to the corresponding AST node and SQL clause;
Pair Analyzer features such as \emph{Additional Grounding Check}
and \emph{Pattern-Based Business Meaning Support} map to canonical
rule statuses and their supporting explanations; and features such as
\emph{Intent Alignment: Grouping} and
\emph{Question Planner: Output Structure} map to probe verdicts,
supporting evidence, rationales, and associated diagnostic records.

This provides a direct provenance path from an influential model
feature to the diagnostic condition from which it was derived.
A representative candidate-level prediction trace illustrating this
feature-to-evidence connection for a false-positive verification decision is
provided in the supplementary material.

\subsection{Verification on Ground-Truth SQL}

As a complementary evaluation, we apply both verifiers to 1,534 ground-truth
SQL queries from the BIRD development databases. Unlike the generated-SQL
evaluation, this setting contains only the reference SQL associated with each
question and therefore contains no negative candidates. It measures how often
each verifier assigns a positive decision when evaluated on the available
ground-truth SQL. Since all instances belong to the positive class, accuracy is equivalent to recall, precision is 100\% by construction,
and ROC-AUC is undefined.

As shown in Table~\ref{tab:ground-overall}, GradeSQL-7B accepts 62.84\% of the ground-truth SQL queries,
compared with 60.17\% for TraceSQL. At the
database level, TraceSQL achieves higher acceptance on California Schools,
Codebase Community, Financial, Formula 1, and Thrombosis Prediction, while
GradeSQL-7B is higher on the remaining databases. Complete database-level
results are provided in the supplementary material.

We use this experiment as a complementary evaluation of verifier behavior on
ground-truth SQL, while the generated-SQL setting remains the primary
evaluation for comparing binary verification performance.

\begin{table}[!tb]
\caption{Verification performance on 1,534 ground-truth SQL queries from the
BIRD development databases. Since all instances belong to the positive class,
Accuracy equals Recall, Precision is 100\%, and ROC-AUC is undefined.}
\label{tab:ground-overall}
\centering
\small

\begin{tabular}{lcccc}
\toprule
Model & Acc./Rec. & Prec. & F1 & AUC \\
\midrule
GradeSQL-7B
& \textbf{62.84}
& 100.00
& \textbf{77.18}
& N/A \\

TraceSQL
& 60.17
& 100.00
& 75.13
& N/A \\

\bottomrule
\end{tabular}
\end{table}

\section{Discussion}

\subsection{Answerability Through Diagnostic Verification Evidence}

TraceSQL and GradeSQL use the same type of execution-derived candidate
supervision, but TraceSQL exposes a structured diagnostic representation at
inference time rather than only a final correctness score. This representation
combines question-level signals, such as ambiguity and planning requirements,
with candidate-level signals covering schema grounding, SQL structure, and
intent alignment.

This distinction is useful because verification depends on both the question
and the candidate SQL . A clearly specified question can still be paired with
a candidate that does not realize the intended request, while ambiguity in the
question can make the reliability of an otherwise valid candidate SQL harder
to assess. TraceSQL captures these complementary signals explicitly and
uses them to support reference-free verification.

\subsection{Traceability and Generality}

The feature analysis shows that the learned verifier relies on multiple forms
of evidence. Deterministic SQL-structure features such as
\emph{SQL DISTINCT Usage}, \emph{SQL LIMIT Usage}, and
\emph{SQL Aggregation Usage} appear alongside grounding and semantic-validity signals from the Pair
Analysis family, 
such as \emph{Additional Grounding Check},
\emph{Pattern-Based Business Meaning Support}, and
\emph{Data-Validity Rule Support}. This suggests that the model combines
information about what operations are present in the SQL with evidence about
whether those operations are supported by the question and database context.

Because the features have fixed semantic definitions and remain linked to their
underlying diagnostic records, influential predictions can be inspected and
traced back to concrete evidence. The full diagnostic vocabulary is retained in the representation, allowing
signals that are weak on the current BIRD distribution to remain available
under different database and query settings.

\section{Future Work}
A primary direction for future work is to scale TraceSQL training from the
current 2,000-candidate controlled sample to the complete 5,284-candidate
paired source while retaining the same held-out 11-database evaluation. This
will allow us to examine whether additional supervision improves verification
performance and whether the learned importance of the diagnostic signals
remains stable as the training set expands.

We also plan to investigate a tighter integration between TraceSQL and
GradeSQL-style outcome reward modeling. Rather than using the diagnostic
representation as a separate verifier, the 67 TraceSQL features can be
incorporated directly into the GradeSQL verification model as additional
structured evidence. This would enable a controlled study of how augmenting an ORM with explicit
ambiguity, planning, grounding, SQL-structure, and intent-alignment signals
affects predictive performance relative to the original GradeSQL formulation.

This comparison will help clarify whether the TraceSQL representation is more
useful as a standalone verifier or as an additional evidence layer within a
larger learned verification model.

\section{Conclusion}

We presented TraceSQL, a reference-free Text-to-SQL verifier built around
traceable diagnostic evidence. TraceSQL constructs a 67-feature diagnostic
representation covering ambiguity, question requirements,
question--schema--SQL consistency, SQL structure, and intent
alignment, while retaining the underlying diagnostic evidence for further
inspection.

On held-out BIRD development databases, TraceSQL improves over GradeSQL-7B,
reaching 66.47\% F1 and 64.48\% ROC-AUC compared with 61.87\% and 58.26\%,
respectively. The feature analysis also shows that the verifier draws on both
deterministic SQL-structure signals and semantic grounding evidence, suggesting
that reliable verification benefits from combining multiple forms of
diagnostic information.

A key property of TraceSQL is that the verification process remains
inspectable. Influential features can be connected to meaningful diagnostic conditions and
traced back to the evidence from which they were derived. The results show that
reference-free Text-to-SQL verification can be learned from structured
diagnostic signals while preserving evidence that supports inspection and
failure analysis.


\bibliographystyle{ACM-Reference-Format}
\bibliography{tracesql_references}


\appendix

\begin{center}
    {\LARGE\bfseries Supplementary Material}
\end{center}
\vspace{1em}
\section{FLAML Preprocessing Details}
\label{supp:flaml_config}

The main paper reports the FLAML training and model-selection configuration.
For reproducibility, we additionally record the five input features that are
not retained after FLAML preprocessing and feature transformation. All five
belong to the Pair Analysis family. Their absence from the transformed
representation reflects FLAML preprocessing rather than an explicit
feature-selection procedure.

\begin{table}[!b]
\centering
\caption{Pair Analysis input features not retained after FLAML preprocessing
and feature transformation.}
\label{tab:supp_removed_features}
\small
\begin{tabularx}{\linewidth}{@{}X@{}}
\toprule
\textbf{Feature} \\
\midrule
Versioning Rule Support \\
Name and Alias Support \\
Effective Date Meaning \\
User Context Mapping Support \\
User Context Resolution Support \\
\bottomrule
\end{tabularx}
\end{table}

\section{Complete Feature Definitions}
\label{supp:feature_definitions}

The unified representation contains 67 candidate-level features from five
families: 10 ambiguity features, 5 question-planning features, 32 pair-analysis
features, 10 deterministic SQL-structure features, and 10 intent-alignment
features. Richer evidence---including explanations, rationales, confidence,
warnings, and SQL fragments---is retained separately for traceability and is
not included as additional model inputs.
Canonical field names are shown in monospaced type where applicable;
the main paper uses shortened display labels for readability.

\begin{table}[t]
\centering
\caption{TraceSQL feature families.}
\label{tab:supp_feature_families}
\small
\begin{tabularx}{\linewidth}{@{}l c X@{}}
\toprule
\textbf{Feature family} & \textbf{Count} & \textbf{Extraction} \\
\midrule
Ambiguity & 10 & Question-level ambiguity status, probability, and eight probe verdicts. \\
Question Planning & 5 & Probe verdicts comparing planned question requirements with candidate SQL support. \\
Pair Analysis & 32 & Direct projection of canonical Pair Analyzer rule statuses. \\
SQL Structure & 10 & Deterministic binary indicators extracted from the SQL AST. \\
Intent Alignment & 10 & Four evaluator fields and six probe-based alignment verdicts. \\
\midrule
Total & 67 & Fixed candidate-level representation supplied to FLAML. \\
\bottomrule
\end{tabularx}
\end{table}

\subsection{Ambiguity Features}
\label{supp:ambiguity_features}

The ten ambiguity features are summarized in
Table~\ref{tab:supp_ambiguity_features}. Ambiguity is assessed once per
question--database pair and shared across candidate SQL queries for the same
question.

\begin{table*}[!t]
\centering
\caption{Ambiguity features used in TraceSQL.}
\label{tab:supp_ambiguity_features}
\small
\begin{tabularx}{\textwidth}{@{}l l X@{}}
\toprule
\textbf{Probe ID} & \textbf{Feature Name} & \textbf{Meaning} \\
\midrule
-- &
\texttt{Ambiguity\_Overall\_Status} &
Overall Boolean assessment indicating whether the user question is considered
ambiguous. \\

-- &
\texttt{Ambiguity\_Assessment\_Probability} &
Estimated degree of ambiguity assigned by the Ambiguity Detector on a
0--100 scale. \\

AQ001 &
\texttt{Ambiguity\_Metric\_Clarity} &
Captures uncertainty concerning the metric, measure, quantitative value, or
derived calculation requested by the question. \\

AQ002 &
\texttt{Ambiguity\_Entity\_Clarity} &
Captures uncertainty concerning the entity, subject, record, or target
referenced by the question. \\

AQ003 &
\texttt{Ambiguity\_Time\_Scope\_Clarity} &
Captures uncertainty concerning the time period or temporal scope required by
the question. \\

AQ004 &
\texttt{Ambiguity\_Constraint\_Clarity} &
Captures uncertainty concerning filters, thresholds, limits, or other logical
constraints expressed in the question. \\

AQ005 &
\texttt{Ambiguity\_Interpretation\_Clarity} &
Captures whether the question admits multiple plausible semantic
interpretations. \\

AQ006 &
\texttt{Ambiguity\_Reference\_Resolution\_Clarity} &
Captures unresolved references, antecedents, pronouns, or other implicit
references in the question. \\

AQ007 &
\texttt{Ambiguity\_External\_Context\_Clarity} &
Captures whether interpretation of the question depends on unavailable
business knowledge, domain conventions, or external context. \\

AQ008 &
\texttt{Ambiguity\_Analytical\_Intent\_Clarity} &
Captures uncertainty concerning the analytical objective or operation requested
by the user. \\
\bottomrule
\end{tabularx}
\end{table*}

\subsection{Question-Planning Features}
\label{supp:question_planning_features}

The five question-planning features are summarized in
Table~\ref{tab:supp_question_planning_features}.

\begin{table*}[!t]
\centering
\caption{Question-planning features used in TraceSQL.}
\label{tab:supp_question_planning_features}
\small
\begin{tabularx}{\textwidth}{@{}l l X@{}}
\toprule
\textbf{Probe ID} & \textbf{Feature Name} & \textbf{Meaning} \\
\midrule
QP001 &
\texttt{Question\_Planner\_Analytical\_Operation\_Support} &
Whether the candidate SQL supports the analytical operation required by the
question, such as lookup, aggregation, comparison, ranking, or trend analysis. \\

QP002 &
\texttt{Question\_Planner\_Output\_Structure\_Support} &
Whether the candidate SQL supports the expected output columns, answer
structure, and result granularity required by the question. \\

QP003 &
\texttt{Question\_Planner\_Grouping\_Support} &
Whether the candidate SQL supports the grouping requirements and aggregation
dimensions required by the question. \\

QP004 &
\texttt{Question\_Planner\_Ordering\_or\_Ranking\_Support} &
Whether the candidate SQL supports ordering, ranking, top-$k$, sort direction,
or extremum requirements expressed by the question. \\

QP005 &
\texttt{Question\_Planner\_SQL\_Operations\_Support} &
Whether the candidate SQL contains the operations required to answer the
question, such as joins, filters, aggregation, window functions,
\texttt{DISTINCT}, or set operations. \\
\bottomrule
\end{tabularx}
\end{table*}

\subsection{Pair-Analysis Features}
\label{supp:pair_analysis_features}

The 32 pair-analysis features are summarized in
Tables~\ref{tab:supp_pair_features_1} and~\ref{tab:supp_pair_features_2}.
The model receives the canonical rule status for each feature. Rule scores and
explanations are retained as diagnostic evidence but are not additional model
features.

\begin{table*}[!t]
\centering
\caption{Pair-analysis features used in TraceSQL (Part I).}
\label{tab:supp_pair_features_1}
\scriptsize
\begin{tabularx}{\textwidth}{@{}
    >{\raggedright\arraybackslash}p{3.3cm}
    >{\raggedright\arraybackslash}p{4.8cm}
    X@{}}
\toprule
\textbf{Rule ID} & \textbf{Feature Name} & \textbf{Meaning} \\
\midrule
\makecell[l]{\texttt{schema\_reference}\\\texttt{\_verification}} &
\texttt{Pair\_Analyzer\_Schema\_Reference\_Support} &
Whether referenced schemas are present and supported by the catalog. \\

\makecell[l]{\texttt{table\_reference}\\\texttt{\_verification}} &
\texttt{Pair\_Analyzer\_Table\_Reference\_Support} &
Whether referenced tables are present and supported by the catalog. \\

\makecell[l]{\texttt{column\_reference}\\\texttt{\_verification}} &
\texttt{Pair\_Analyzer\_Column\_Reference\_Support} &
Whether referenced columns are present and supported by the catalog. \\

\makecell[l]{\texttt{schema\_scope\_description}\\\texttt{\_verification}} &
\texttt{Pair\_Analyzer\_Schema\_Scope\_Support} &
Whether schema-level descriptions support the question and SQL intent. \\

\makecell[l]{\texttt{table\_scope\_description}\\\texttt{\_verification}} &
\texttt{Pair\_Analyzer\_Table\_Scope\_Support} &
Whether table descriptions support the question and SQL intent. \\

\makecell[l]{\texttt{column\_scope\_description}\\\texttt{\_verification}} &
\texttt{Pair\_Analyzer\_Column\_Scope\_Support} &
Whether column descriptions support the question and SQL intent. \\

\makecell[l]{\texttt{join\_relationship}\\\texttt{\_verification}} &
\texttt{Pair\_Analyzer\_Join\_Relationship\_Support} &
Whether the join relationship is explicitly supported by catalog metadata. \\

\makecell[l]{\texttt{join\_key}\\\texttt{\_verification}} &
\texttt{Pair\_Analyzer\_Join\_Key\_Support} &
Whether the join keys are explicitly documented as relationship keys. \\

\makecell[l]{\texttt{versioning\_rule}\\\texttt{\_verification}} &
\texttt{Pair\_Analyzer\_Versioning\_Rule\_Support} &
Whether required versioning, current-record, effective-dating, or snapshot
rules are documented. \\

\makecell[l]{\texttt{sql\_identifier\_alignment}\\\texttt{\_verification}} &
\texttt{Pair\_Analyzer\_SQL\_Name\_Alignment} &
Whether SQL identifiers match catalog names or documented aliases. \\

\makecell[l]{\texttt{naming\_alias}\\\texttt{\_verification}} &
\texttt{Pair\_Analyzer\_Name\_and\_Alias\_Support} &
Whether abbreviations, prefixes, suffixes, or aliases used in SQL are
documented. \\

\makecell[l]{\texttt{identifier\_semantics}\\\texttt{\_verification}} &
\texttt{Pair\_Analyzer\_Identifier\_Meaning\_Support} &
Whether identifier columns used in joins or filters have documented business
meaning. \\

\makecell[l]{\texttt{numeric\_operation}\\\texttt{\_datatype\_support\_verification}} &
\texttt{Pair\_Analyzer\_Numeric\_Data\_Type\_Support} &
Whether numeric operations use columns with compatible data types. \\

\makecell[l]{\texttt{join\_column\_datatype}\\\texttt{\_compatibility\_verification}} &
\texttt{Pair\_Analyzer\_Join\_Data\_Type\_Compatibility} &
Whether columns used in joins have compatible data types. \\

\makecell[l]{\texttt{sql\_datatype\_support}\\\texttt{\_verification}} &
\texttt{Pair\_Analyzer\_SQL\_Data\_Type\_Support} &
Whether data-type metadata is sufficient to validate SQL behavior. \\

\makecell[l]{\texttt{temporal\_semantics}\\\texttt{\_verification}} &
\texttt{Pair\_Analyzer\_Date\_Time\_Meaning\_Support} &
Whether date/time columns have sufficient documented business meaning to
validate how the SQL uses them. \\
\bottomrule
\end{tabularx}
\end{table*}

\begin{table*}[!t]
\centering
\caption{Pair-analysis features used in TraceSQL (Part II).}
\label{tab:supp_pair_features_2}
\scriptsize
\begin{tabularx}{\textwidth}{@{}
    >{\raggedright\arraybackslash}p{3.3cm}
    >{\raggedright\arraybackslash}p{4.8cm}
    X@{}}
\toprule
\textbf{Rule ID} & \textbf{Feature Name} & \textbf{Meaning} \\
\midrule
\makecell[l]{\texttt{temporal\_fields}\\\texttt{\_verification}} &
\texttt{Pair\_Analyzer\_Date\_Time\_Field\_Distinction} &
Whether similar date/time fields are clearly distinguished so that the
intended field can be identified. \\

\makecell[l]{\texttt{temporal\_queries}\\\texttt{\_verification}} &
\texttt{Pair\_Analyzer\_Temporal\_Analysis\_Support} &
Whether temporal metadata supports historical, period-based, fiscal, or
as-of analysis requested by the question. \\

\makecell[l]{\texttt{timezone\_semantics}\\\texttt{\_verification}} &
\texttt{Pair\_Analyzer\_Timezone\_Handling\_Support} &
Whether timezone handling is documented sufficiently to interpret operational
timestamps correctly. \\

\makecell[l]{\texttt{effective\_date\_semantics}\\\texttt{\_verification}} &
\texttt{Pair\_Analyzer\_Effective\_Date\_Meaning} &
Whether effective dates and validity periods clearly indicate when records
become active, expire, or apply. \\

\makecell[l]{\texttt{metric\_definition}\\\texttt{\_verification}} &
\texttt{Pair\_Analyzer\_Metric\_Definition\_Support} &
Whether a metric is documented sufficiently to validate the metric requested
by the question. \\

\makecell[l]{\texttt{metric\_aggregation\_logic}\\\texttt{\_verification}} &
\texttt{Pair\_Analyzer\_Metric\_Calculation\_Support} &
Whether aggregations, ratios, and rollups follow the documented meaning of the
metric. \\

\makecell[l]{\texttt{table\_role\_semantics}\\\texttt{\_verification}} &
\texttt{Pair\_Analyzer\_Table\_Role\_Clarity} &
Whether the catalog identifies the role of the table, such as master,
transaction, history, snapshot, summary, or bridge. \\

\makecell[l]{\texttt{null\_semantics}\\\texttt{\_verification}} &
\texttt{Pair\_Analyzer\_NULL\_Handling\_Support} &
Whether SQL handling of \texttt{NULL} values is consistent with documented
NULL behavior. \\

\makecell[l]{\texttt{data\_validity\_rule}\\\texttt{\_verification}} &
\texttt{Pair\_Analyzer\_Data\_Validity\_Rule\_Support} &
Whether predicates such as active, current, enabled, effective, latest, or
approved are supported by documented business rules. \\

\makecell[l]{\texttt{pattern\_based\_business}\\\texttt{\_meaning\_verification}} &
\texttt{Pair\_Analyzer\_Pattern\_Based\_Meaning\_Support} &
Whether \texttt{LIKE}, regular-expression, substring, prefix, or encoded-code
logic has documented business meaning. \\

\makecell[l]{\texttt{pattern\_mapping}\\\texttt{\_verification}} &
\texttt{Pair\_Analyzer\_Pattern\_Mapping\_Support} &
Whether patterns used by the SQL map to documented business categories or
values. \\

\makecell[l]{\texttt{business\_term\_mapping}\\\texttt{\_verification}} &
\texttt{Pair\_Analyzer\_Business\_Term\_Mapping\_Support} &
Whether business terms in the question map to documented catalog entities,
values, or terminology. \\

\makecell[l]{\texttt{alias\_or\_canonical}\\\texttt{\_mapping\_verification}} &
\texttt{Pair\_Analyzer\_Canonical\_Name\_Mapping\_Support} &
Whether aliases or alternate names map to documented canonical schema names
or synonyms. \\

\makecell[l]{\texttt{user\_context\_mapping}\\\texttt{\_verification}} &
\texttt{Pair\_Analyzer\_User\_Context\_Mapping\_Support} &
Whether user-context expressions such as ``my'', ``our'', or ``assigned to
me'' map to documented business concepts. \\

\makecell[l]{\texttt{user\_context\_resolution}\\\texttt{\_verification}} &
\texttt{Pair\_Analyzer\_User\_Context\_Resolution\_Support} &
Whether a user-context expression can be resolved to a documented catalog
meaning or ownership relationship. \\

\makecell[l]{\texttt{llm\_detectable\_pairwise}\\\texttt{\_gap\_verification}} &
\texttt{Pair\_Analyzer\_Additional\_Grounding\_Check} &
Whether any additional catalog or semantic grounding issue materially affects
the reliability of the question and candidate SQL pair. \\
\bottomrule
\end{tabularx}
\end{table*}

\subsection{SQL-Structure Features}
\label{supp:sql_structure_features}

The candidate SQL is parsed using \texttt{SQLGlot}, and the resulting AST is
inspected deterministically. The ten binary structure indicators are listed in
Table~\ref{tab:supp_sql_structure_features}. Parse failures remain diagnostic
artifacts rather than being silently interpreted as a particular SQL structure.

\begin{table*}[!t]
\centering
\caption{Deterministic SQL-structure features used in TraceSQL.}
\label{tab:supp_sql_structure_features}
\small
\begin{tabularx}{\textwidth}{@{}lX@{}}
\toprule
\textbf{Feature Name} & \textbf{Meaning} \\
\midrule
SQL Aggregation Usage & Whether the candidate contains an aggregate operation. \\
SQL Grouping Usage & Whether the candidate contains a \texttt{GROUP BY} operation. \\
SQL Join Usage & Whether the candidate joins multiple relations. \\
SQL Filtering Usage & Whether the candidate contains \texttt{WHERE} or \texttt{HAVING} filtering. \\
SQL Temporal Filter Usage & Whether the candidate contains a temporal/date-related filter. \\
SQL ORDER BY Usage & Whether the candidate contains an \texttt{ORDER BY} clause. \\
SQL LIMIT Usage & Whether the candidate contains a \texttt{LIMIT} clause. \\
SQL Subquery Usage & Whether the candidate contains a subquery or common table expression. \\
SQL Window Function Usage & Whether the candidate contains a window-function expression. \\
SQL DISTINCT Usage & Whether the candidate requests \texttt{DISTINCT} results. \\
\bottomrule
\end{tabularx}
\end{table*}

\subsection{Intent-Alignment Features}
\label{supp:intent_alignment_features}

Intent alignment combines four direct evaluator fields with six probe-based
alignment verdicts. The ten resulting features are summarized in
Table~\ref{tab:supp_intent_alignment_features}.

\begin{table*}[!tb]
\centering
\caption{Intent-alignment features used in TraceSQL.}
\label{tab:supp_intent_alignment_features}
\small
\begin{tabularx}{\textwidth}{@{}lX@{}}
\toprule
\textbf{Feature Name} & \textbf{Meaning} \\
\midrule
Intent Alignment Overall Score & Reference-free evaluator score for overall alignment between the question and candidate SQL. \\
Intent Alignment Confidence & Confidence associated with the reference-free alignment evaluation. \\
Intent Alignment Gate Decision & Evaluator gate decision, such as accepting or requesting intervention. \\
Intent Alignment Intervention Reason & Categorized reason for intervention when the evaluator does not accept the candidate. \\
Intent Alignment: Requested Metric & Whether the metric or measure requested by the question is aligned with the candidate SQL. \\
Intent Alignment: Filter Conditions & Whether filtering constraints in the question are represented by the candidate SQL. \\
Intent Alignment: Entity or Scope & Whether the candidate SQL matches the requested entity and analytical scope. \\
Intent Alignment: Grouping & Whether grouping requirements in the question align with the candidate SQL. \\
Intent Alignment: Ranking or Extremum & Whether ranking, top-$k$, minimum, maximum, or related requirements align with the candidate SQL. \\
Intent Alignment: Business-Term Mapping & Whether business terminology in the question is aligned with the schema and SQL interpretation. \\
\bottomrule
\end{tabularx}
\end{table*}

\section{Detailed Feature-Importance Analysis}
\label{supp:feature_analysis}

The main paper reports the leading permutation-importance and SHAP results.
Here we provide the global SHAP ranking and a cross-method comparison with
permutation importance and native Extra Trees feature importance. These
analyses characterize model reliance on the training/model-development
population.

\subsection{Global SHAP Analysis}
\label{supp:shap_analysis}

\begin{figure*}[!t]
    \centering
    \includegraphics[width=0.90\textwidth]{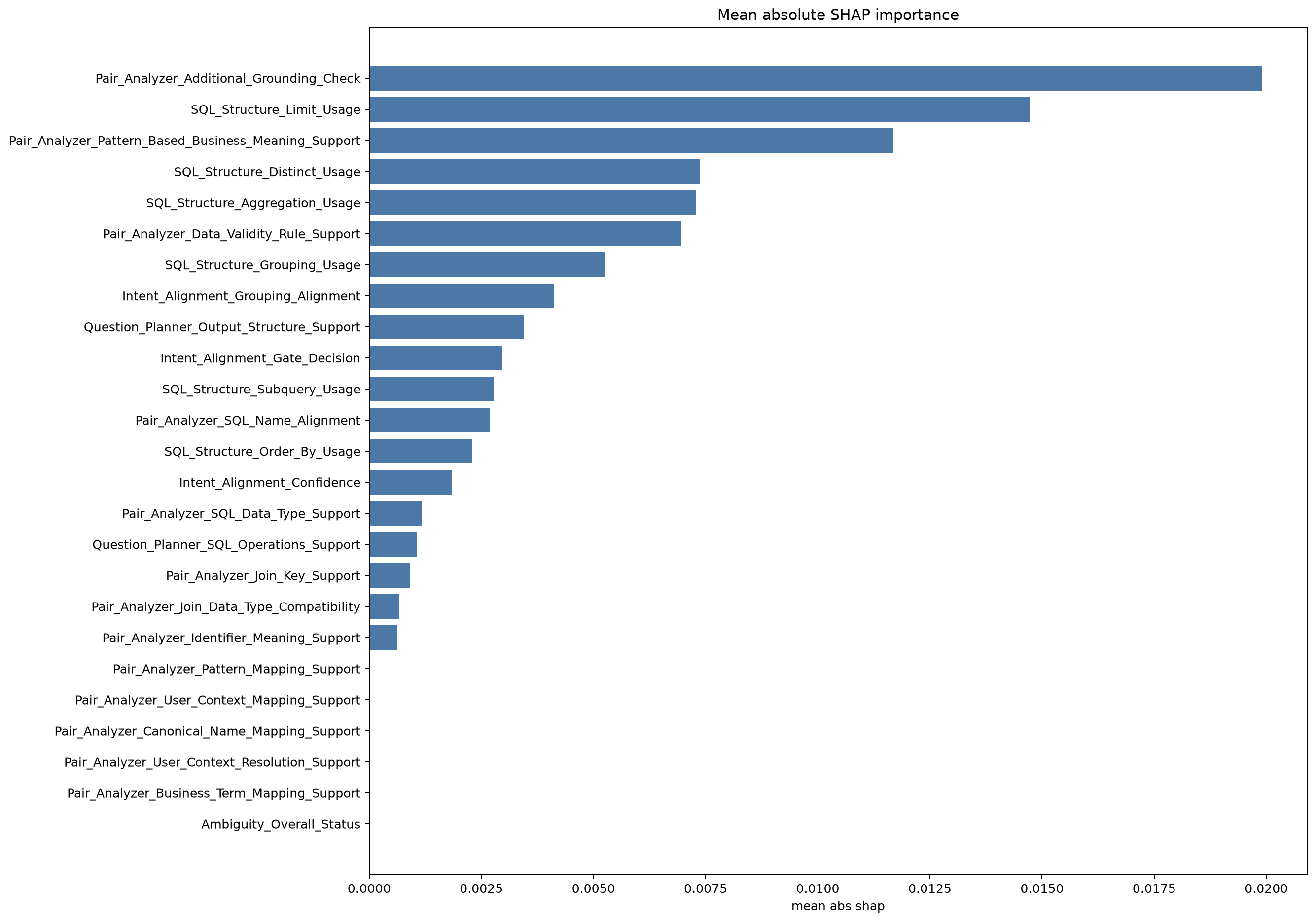}
    \caption{Mean absolute SHAP importance for the fifteen most influential
    TraceSQL features. Larger values indicate greater average contribution to
    the fitted model output.}
    \Description{Horizontal bar chart showing the fifteen TraceSQL features
    with the largest mean absolute SHAP values.}
    \label{fig:supp_shap_mean_abs}
\end{figure*}

Figure~\ref{fig:supp_shap_mean_abs} reports mean absolute SHAP values for the
fifteen highest-ranked features.

The global SHAP ranking is led by \emph{Additional Grounding Check}
(\(0.0199\)), followed by \emph{SQL LIMIT Usage} (\(0.0147\)) and
\emph{Pattern-Based Business Meaning Support} (\(0.0117\)).
\emph{SQL DISTINCT Usage} (\(0.0074\)), \emph{SQL Aggregation Usage}
(\(0.0073\)), and \emph{Data-Validity Rule Support} (\(0.0070\)) form the
next group. The leading signals therefore include both deterministic
SQL-structure properties and semantic grounding evidence.

A second group includes \emph{SQL Grouping Usage} (\(0.0052\)),
\emph{Intent Alignment: Grouping} (\(0.0041\)), and
\emph{Question Planner: Output Structure} (\(0.0034\)). The remaining
top-fifteen features have smaller mean absolute contributions.

\begin{table}[!t]
\centering
\caption{Cross-method agreement among the leading TraceSQL features.
Lower rank indicates greater importance within each method.}
\label{tab:supp_importance_agreement}
\small
\setlength{\tabcolsep}{5pt}
\renewcommand{\arraystretch}{1.05}

\begin{tabular}{@{}lccc@{}}
\toprule
\textbf{Feature} &
\textbf{SHAP} &
\textbf{Perm.} &
\textbf{Native} \\
\midrule
Additional Grounding Check & 1 & 4 & 2 \\
SQL LIMIT Usage & 2 & 2 & 5 \\
Pattern-Based Business Meaning Support & 3 & 5 & 3 \\
SQL DISTINCT Usage & 4 & 1 & 14 \\
SQL Aggregation Usage & 5 & 3 & 1 \\
Data-Validity Rule Support & 6 & 6 & 4 \\
SQL Grouping Usage & 7 & 9 & 8 \\
Intent Alignment: Grouping & 8 & 7 & 17 \\
Question Planner: Output Structure & 9 & 8 & 12 \\
SQL Subquery Usage & 11 & 10 & 16 \\
\bottomrule
\end{tabular}
\end{table}

\subsection{Coverage Across Diagnostic Feature Families}
\label{supp:family_coverage}

The leading features are drawn from multiple parts of the TraceSQL
representation rather than from a single diagnostic family.
Table~\ref{tab:supp_family_coverage} summarizes the number of features from
each family appearing in the top ten under SHAP, permutation importance, and
native Extra Trees feature importance. This analysis describes representation
coverage among highly ranked signals; it should not be interpreted as a
feature-family ablation or as a measure of the causal contribution of each
family.

\begin{table}[!tb]
\centering
\caption{Feature-family representation among the top ten features under each
importance method.}
\label{tab:supp_family_coverage}
\small
\begin{tabular}{lccc}
\toprule
\textbf{Feature Family} &
\textbf{SHAP} &
\textbf{Perm.} &
\textbf{Native} \\
\midrule
Ambiguity         & 0 & 0 & 0 \\
Question Planning & 1 & 1 & 1 \\
Pair Analysis     & 3 & 3 & 4 \\
SQL Structure     & 4 & 5 & 3 \\
Intent Alignment  & 2 & 1 & 2 \\
\bottomrule
\end{tabular}
\end{table}

Across all three methods, SQL Structure and Pair Analysis account for most of
the highly ranked features, while Question Planning and Intent Alignment also
contribute signals to the leading set. Importantly, the rankings do not reduce
to SQL structure alone: semantic grounding and validity signals from the Pair
Analysis family consistently appear among the strongest features.

\subsection{Cross-Method Feature-Importance Comparison}
\label{supp:complementary_importance}

Permutation importance, SHAP, and native Extra Trees feature importance
measure different aspects of model reliance. Table~\ref{tab:supp_importance_agreement}
therefore compares ranks rather than placing their raw values on a common
scale.

The strongest cross-method agreement is observed for
\emph{Additional Grounding Check}, \emph{SQL LIMIT Usage},
\emph{Pattern-Based Business Meaning Support},
\emph{SQL Aggregation Usage}, and \emph{Data-Validity Rule Support}, all of
which rank near the top under SHAP, permutation importance, and native feature
importance. \emph{SQL DISTINCT Usage} is more method-sensitive: it ranks first
under permutation importance and fourth under SHAP, but lower under native
feature importance. Overall, the recurring high-ranked set spans both
SQL-structure and semantic diagnostic signals. Ambiguity features do not appear within the top ten under the three importance
methods. In the current training, ambiguity is assessed at the
question--database level and is therefore shared across alternative candidate
SQLs associated with the same question. In contrast, SQL-structure, grounding,
and intent-alignment features can vary across candidates. This difference may
reduce the ability of ambiguity features to distinguish positive and negative
candidate SQLs within the paired training setting. The result should therefore
be interpreted as limited model reliance on the current ambiguity
representation under this training distribution rather than as evidence that
question ambiguity is uninformative for Text-to-SQL verification.

\begin{figure*}[!t]
    \centering
    \includegraphics[width=0.82\textwidth]
    {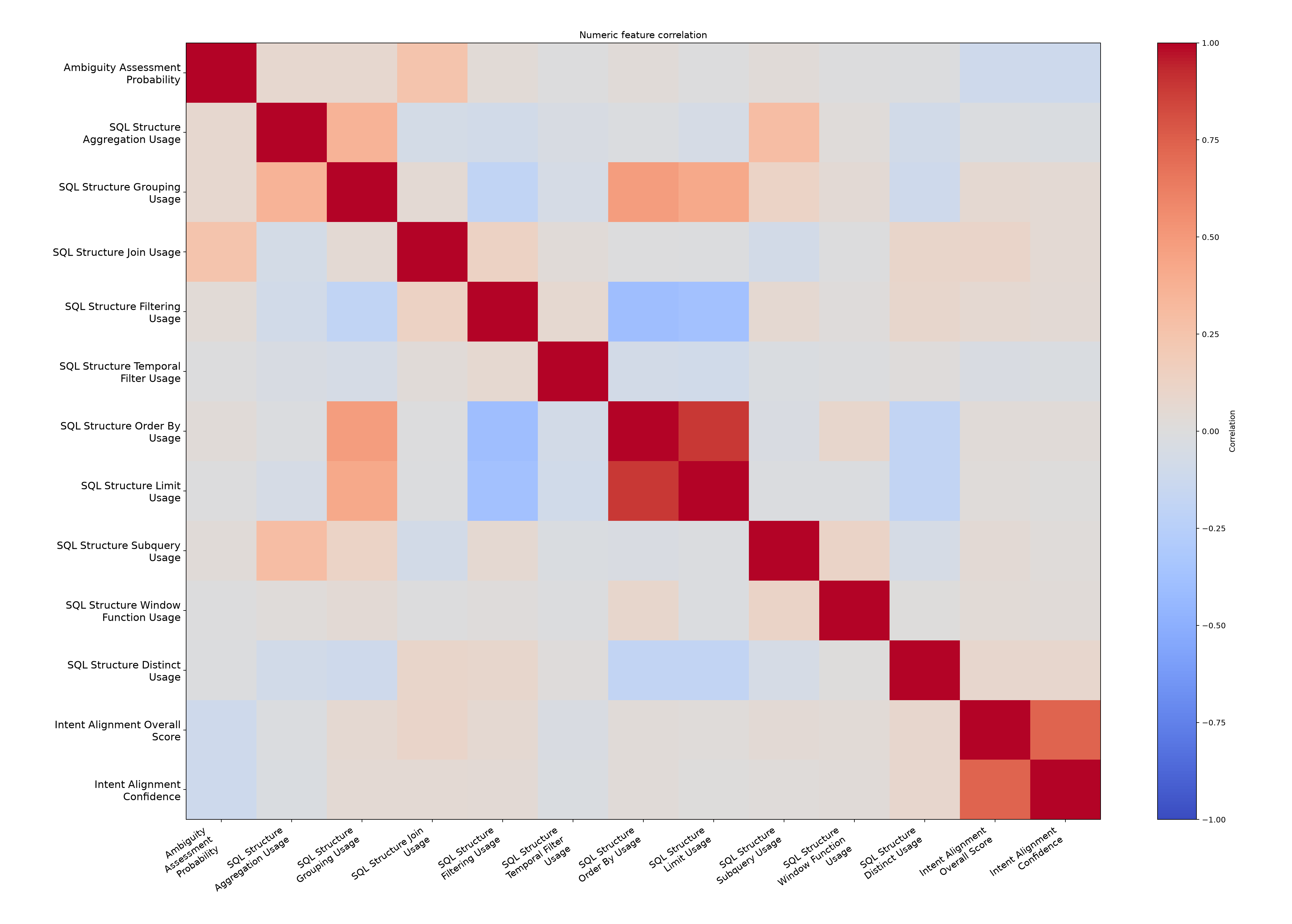}
    \caption{Pairwise correlation matrix for the numeric TraceSQL features.}
    \Description{Correlation heatmap for ambiguity probability, SQL-structure
    indicators, and intent-alignment score and confidence.}
    \label{fig:supp_numeric_corr}
\end{figure*}

\begin{figure*}[!t]
    \centering
    \includegraphics[width=0.92\textwidth]{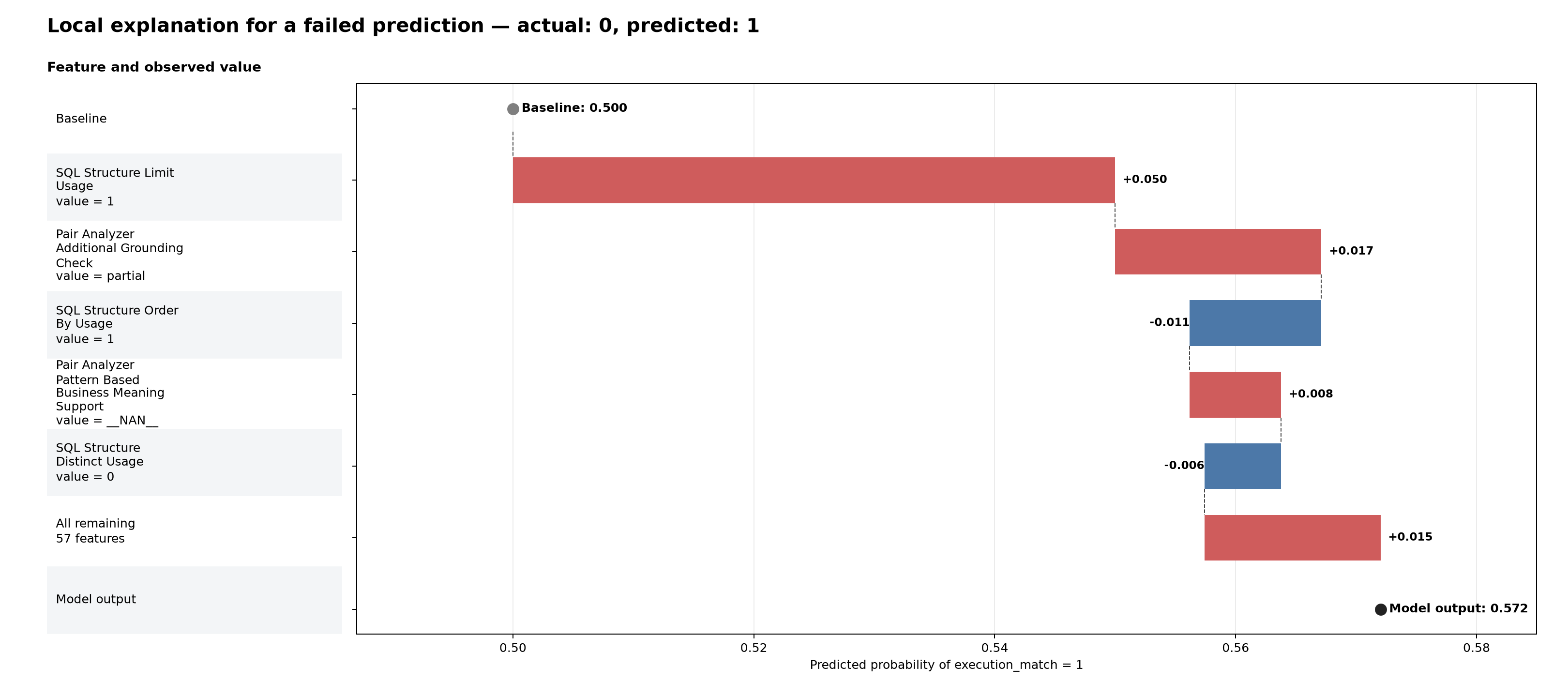}
    \caption{Local SHAP explanation for the representative false-positive
    verification decision.}
    \Description{Local feature-attribution plot for a false-positive TraceSQL
    prediction on a Formula 1 question about the 2017 Chinese Grand Prix.}
    \label{fig:supp_representative_trace}
\end{figure*}

\section{Feature Correlation Analysis}
\label{supp:feature_corr}

We examine pairwise correlations among numeric TraceSQL features to identify
strongly related signals that may contain overlapping information.

Most numeric feature pairs exhibit weak linear relationships. Two notable
dependencies are \emph{SQL ORDER BY Usage} and \emph{SQL LIMIT Usage}
(\(r=0.887\)), and \emph{Intent Alignment Overall Score} and
\emph{Intent Alignment Confidence} (\(r=0.729\)). These dependencies provide
context for differences across feature-importance methods, since correlated
features can share or redistribute predictive information. The correlations
are descriptive of the training population.

\section{Representative Prediction Trace}
\label{supp:traceability_example}

To illustrate candidate-level traceability, we examine a representative
false-positive case from the generated-SQL evaluation. The question asks for
the top three drivers and their points in the 2017 Chinese Grand Prix. The
generated candidate SQL filters the relevant race, joins the race, results,
and driver tables, orders the candidates by points, and applies
\texttt{LIMIT 3}. The execution-match label is negative, while TraceSQL
assigns a positive probability of \(0.572\).

The retained diagnostic evidence makes the source of this prediction
inspectable. The candidate SQL contains structural signals that are globally
influential in the fitted model, including ordering, \texttt{LIMIT}, filtering,
and joins. Pair Analyzer diagnostics support the referenced tables, columns,
and join relationship, while the intent-alignment diagnostics mark ranking and
business-term alignment as \textit{PASS}, with an overall intent score of
\(1.0\) and an \textit{ACCEPT} gate decision. Multiple diagnostic signals
therefore describe the candidate as plausible even though its execution-match
label is negative.

This case illustrates the feature-to-evidence path used for traceability:
local attribution identifies influential model features, and those features
remain linked to named SQL-structure properties, Pair Analyzer rule results,
or probe-level diagnostic records retained during feature extraction.

\section{Additional Database-Level Results}
\label{supp:ground_truth_db}

The main paper reports the aggregate ground-truth SQL evaluation.
Because all 1,534 instances belong to the positive class, accuracy equals
recall, precision is 100\%, and ROC-AUC is undefined. We therefore report
database-level acceptance rates in Table~\ref{tab:supp_ground_db}.

\begin{table}[!tb]
\centering
\caption{Database-level acceptance rate (\%) on the 1,534 ground-truth SQL
queries. Because all instances belong to the positive class, acceptance rate
is equivalent to accuracy and recall. Bold indicates the higher result.}
\label{tab:supp_ground_db}
\small
\setlength{\tabcolsep}{4pt}
\renewcommand{\arraystretch}{1.02}

\begin{tabular}{@{}lcc@{}}
\toprule
\textbf{Database} &
\textbf{GradeSQL-7B} &
\textbf{TraceSQL} \\
\midrule
California Schools        & 41.57 & \textbf{67.42} \\
Card Games                & \textbf{59.16} & 43.98 \\
Codebase Community        & 65.59 & \textbf{66.13} \\
Debit Card Specializing   & \textbf{68.75} & 29.69 \\
European Football 2       & \textbf{65.12} & 62.79 \\
Financial                 & 60.38 & \textbf{75.47} \\
Formula 1                 & 65.52 & \textbf{72.41} \\
Student Club              & \textbf{73.42} & 60.76 \\
Superhero                 & \textbf{79.84} & 65.12 \\
Thrombosis Prediction     & 50.92 & \textbf{54.60} \\
Toxicology                & \textbf{57.93} & 55.86 \\
\midrule
\textbf{Overall}          & \textbf{62.84} & 60.17 \\
\bottomrule
\end{tabular}
\end{table}

TraceSQL has higher acceptance on California Schools, Codebase Community,
Financial, Formula 1, and Thrombosis Prediction, while GradeSQL-7B is higher
on the remaining six databases. The generated-SQL setting remains the primary
two-class verification evaluation.

\end{document}